\documentclass[9pt, conference, a4paper, compsocconf]{IEEEtran}
\usepackage{amsmath,amsxtra,amssymb,amsthm,latexsym,amscd,amsfonts}
\usepackage[utf8]{vietnam}
\usepackage[top=2.7cm, bottom = 5.4cm, left=2.72cm, right = 2.85cm]{geometry}
\usepackage{balance}

\usepackage{fancyhdr}
\usepackage{cite}
\usepackage{algorithmic}
\usepackage{textcomp}
\usepackage{booktabs}   
\usepackage{tabularx}  
\usepackage{xcolor}
\usepackage{amsmath,amsxtra,amssymb,amsthm,latexsym,amscd,amsfonts}
\usepackage{pgfplots}
\usepackage{multirow}
\usepackage{subcaption}
\usepackage{url}
\pgfplotsset{compat=1.17} 

\usepackage{tikz}
\usetikzlibrary{positioning}  
\usetikzlibrary{arrows.meta}  
\usetikzlibrary{shapes}      
\usetikzlibrary{calc}      
\usetikzlibrary{fit}          
\usepackage[english]{babel} 
\usepackage{fancyhdr}
\usepackage{array}

\usepackage{caption}
\usepackage[font=footnotesize]{subfig}

\makeatletter
\def\ps@IEEEtitlepagestyle{%
\def\@oddhead{\hfil \small{\textit{Hội thảo khoa học Quốc gia về Trí tuệ nhân tạo 2026 (FJCAI) - Cần Thơ, 27-28/3/2026}\hfil}%
	\def\@evenhead{\hfil\small{\textit{Hội thảo khoa học quốc gia về Trí tuệ nhân tạo 2026 (FJCAI) - Cần Thơ, 27-28/3/2026}\hfil}}%
		\def\@oddfoot{\scriptsize \thepage \hfil }%
		\def\@evenfoot{\scriptsize \hfil \thepage}
	}
}

\def\@maketitle{%
  \newpage
  \null
  \begin{center}%
    {%
      \fontsize{16}{18}\selectfont
      \bfseries \@title \par
    }%
    \vskip 4em%
    {%
    \fontsize{9}{3.5}\selectfont
      
      \begin{tabular}[t]{c}%
        \@author
      \end{tabular}\par
    }%
  \end{center}%
  \par
}

\renewenvironment{abstract}
  {\normalfont
   \if@twocolumn
     \@IEEEabskeysecsize\bfseries\textit{\abstractname}: 
   \else
     \begin{center}\@IEEEabskeysecsize\textbf{\abstractname}\end{center}
     \@IEEEabskeysecsize
     \setlength{\parindent}{0pt}   
     \noindent
   \fi}
  {\par}

\renewenvironment{abstract}
  {\normalfont
   \@IEEEabskeysecsize
   \setlength{\parindent}{0pt}
   \noindent
   \if@twocolumn
     \bfseries\textit{\abstractname}: 
   \else
     \begin{center}\textbf{\abstractname}\end{center}
   \fi}
  {\par}
  
\renewenvironment{IEEEkeywords}
  {\normalfont
   \@IEEEabskeysecsize
   \vspace{0.3em}       
   \setlength{\parindent}{0pt}
   \noindent
   {\bfseries\itshape \IEEEkeywordsname: }\ignorespaces}
  {\par}

\def\IEEEbibitemsep{1.5pt plus .5pt}

\makeatother

\makeatletter
\AtBeginDocument{
  \renewcommand{\abstractname}{Abstract}
}
\renewcommand{\IEEEkeywordsname}{Keywords}
\makeatother

\begin{document}
\pagenumbering{gobble}
\fontsize{9}{10}
\selectfont

\fancyhead[RE,LO]{\centering{\small{\textit{Hội thảo khoa học Quốc gia về Trí tuệ nhân tạo 2026 (FJCAI) - Cần Thơ, 27-28/3/2026}}}}


%
\title{Diffusion Model in Latent Space for Medical
Image Segmentation Task}


\author{
    \IEEEauthorblockN{
        Ngoc Huynh Trinh,
        Hai Toan Nguyen,
        Son Ba Luong,
        Quoc Long Tran
    }
    \IEEEauthorblockA{
        \textit{Institute for Artificial Intelligence, University of Engineering and Technology},\\
        \textit{Vietnam National University, Hanoi, Vietnam}\\
        Emails: \{huynhtn, nguyenhaitoan, bals, tqlong\}@vnu.edu.vn, 
    }
}

\maketitle
\begin{abstract}
Medical image segmentation is essential for supporting clinical diagnosis and treatment planning. With growing imaging demands, artificial intelligence (AI) models are increasingly used to improve the accuracy and efficiency of this task while reducing the workload of radiologists. However, traditional approaches often produce a single segmentation mask per input image, limiting the ability to capture uncertainty in the segmentation process. Recent advancements in generative models for image generation present new opportunities to address these limitations by enabling the generation of multiple segmentation masks for each input image. Nevertheless, computational complexity and performance remain key challenges when operating directly in the image domain. In this work, we propose MedSegLatDiff, an effective framework for medical image segmentation that integrates a diffusion-based model (DM) with two variational autoencoders (VAEs) to decouple the segmentation process from perceptual data compression. Specifically, VAEs compress data into a low-dimensional latent space, reducing noise and accelerating learning, while the DM performs segmentation more efficiently within this latent representation. Furthermore, we replace the traditional Mean Squared Error (MSE) loss with a weighted Cross-Entropy (WCE) loss in the perceptual mask compression module, which enhances the reconstruction of segmentation masks during the encode–decode process, particularly for tiny nodules. Our designed framework emulates the collaborative segmentation approach of multiple clinicians, resulting in more robust and reliable performance compared to previous methods that model the segmentation patterns of individual practitioners. We evaluate our model on three datasets: ISIC-2018, CVC-Clinic, and LIDC-IDRI, demonstrating competitive performance while simultaneously providing the ability to generate confidence maps for deeper analysis by experts. Our results show that MedSegLatDiff achieves prominent segmentation performance while offering enhanced interpretability and consistency for clinical applications.
\end{abstract}


\IEEEpeerreviewmaketitle

\section{Introduction}
Medical image segmentation, which involves partitioning medical images into meaningful regions, plays a crucial role in clinical applications such as diagnosis, surgical planning, and image-guided procedures~\cite{pham2000current}. Accurate segmentation enables clinicians to better interpret images, facilitate comparisons, and monitor changes over time. However, manually labeling segmentation masks is time-consuming and labor-intensive, motivating the development of automated methods. While classical techniques such as binary thresholding~\cite{senthilkumaran2016image} and k-means clustering~\cite{ng2006medical} have shown limited effectiveness~\cite{jardim2023image, ramesh2021review}, recent advances in AI have enabled the widespread adoption of machine learning (ML) and deep learning (DL) models, which achieve remarkable results~\cite{litjens2017survey}.

Typically, these AI models are trained on pre-labeled datasets and can automatically segment new images. Most existing approaches adopt a one-to-one strategy, producing only a single segmented output per input image~\cite{unet, unet++, nnunet, resunet}. These methods basically learn a mapping function from each input image to a single segmentation mask, thereby replicating the behavior of an individual annotator. Recently, generative models, particularly diffusion models (DMs)~\cite{ddpm}, have led to substantial advances in image synthesis, often outperforming earlier approaches such as Generative Adversarial Networks (GANs)~\cite{gan} and Variational Autoencoders (VAEs)~\cite{vae}, which remain among the most widely used methods in recent years. Beyond their success in natural image generation, these models open up new possibilities for medical image segmentation through a one-to-many paradigm. Unlike the previous one-to-one approach, DMs can generate multiple segmentation masks for a single input, capturing the model's uncertainty in ambiguous medical data. Several recent studies have adopted this strategy and demonstrated superior performance and robustness compared to conventional methods~\cite{ensemblediff, segdiff, medsegdiff, nguyen2025aleatoric}. These results indicate that diffusion-based approaches not only enhance segmentation accuracy but also provide richer information to support clinical decision-making, particularly in challenging cases involving ambiguous or subtle anatomical structures. However, these models typically operate directly in the image space, where performance is limited because they are required to jointly perform image compression and region-of-interest analysis. Consequently, we identify opportunities to further improve model performance by conducting segmentation in a low-dimensional latent space.

In this paper, we propose MedSegLatDiff, a medical image segmentation framework that adopts a novel one-to-many paradigm based on DM. Our framework integrates two VAEs with a diffusion-based model and consists of two main stages. The first stage performs an encoder–decoder architecture to map the input image into a low-dimensional latent representation, while the second stage performs diffusion processes in the latent space to generate a diverse set of segmentation masks for each input image. By conditioning the stochastic generation process of the diffusion model in the latent space, the model produces a set of segmentation masks corresponding to the input, whose variability captures the uncertainty typically exhibited by a group of clinicians. Thus, our model effectively simulates the segmentation consensus among multiple doctors and is well-suited for medical image segmentation tasks. Furthermore, the segmented results from the generated set of masks compete with those of previous models, while confidence maps derived from these outputs assist clinicians in making more thorough and reliable diagnoses.
To the best of our knowledge, the main contributions of this paper are as follows:
\begin{itemize}
\item Training and incorporating VAEs to develop a conditional diffusion-based model in latent space, which reduces noise, compresses images into latent space, and improves both training and inference efficiency.
\item Replacing Mean Squared Error (MSE) with weighted Cross-Entropy (WCE) in the VAE mask reconstruction stage improves the preservation of small structures, especially in cases involving tiny masks, thereby reducing the risk of neglect or misinterpretation as noise.
\item Proposing MedSegLatDiff for medical image segmentation, which simulates the segmentation process of a group of clinicians by generating multiple outputs per input image to approximate their variance. Furthermore, confidence maps derived from these outputs assist experts in performing more thorough and reliable diagnoses.
\end{itemize}
The remainder of the paper is organized as follows. Section~\ref{sec:related_work} reviews recent prominent studies on medical image segmentation, generative models, and autoencoder-based approaches. Section~\ref{sec:method} describes the methods and techniques employed in the proposed model. Section~\ref{sec:experiments} presents experimental results on three datasets to demonstrate the model’s effectiveness. Finally, Section~\ref{sec:conclusion} concludes the study and outlines potential directions for future research.

\section{Related Work} \label{sec:related_work}

\noindent\textbf{Medical Image Segmentation} refers to the process of assigning a label to each pixel (or voxel) in a medical image to determine its class membership, thereby identifying regions of interest with diverse shapes. This problem has been extensively studied using various AI architectures, demonstrating superior performance compared to classical methods~\cite{litjens2017survey}.  Early models followed a one-to-one paradigm, evolving from fully convolutional networks (FCNs)~\cite{fcn}, which employ conventional CNN-based architectures, to encoder–decoder architectures with skip connections, such as U-Net~\cite{unet} and U-Net++~\cite{unet++}, which extend skip connections to improve feature propagation. Traditional CNN-based models, including nnU-Net~\cite{nnunet} and ResUNet~\cite{resunet}, have demonstrated strong performance in medical image segmentation by effectively capturing hierarchical features and leveraging multi-scale contextual information. However, these methods are limited in modeling long-range dependencies, which are essential for capturing global contextual information in complex medical images. To address this limitation, transformer-based architectures such as SegFormer~\cite{segformer}, TransUNet~\cite{transunet}, and SwinUNet~\cite{swinunet} have been introduced to enable more comprehensive image analysis through deep transformer blocks at the cost of increased computational complexity and memory consumption. In addition, hybrid approaches that incorporate hypernetworks, such as HyperSeg~\cite{hyperseg}, have been explored but remain complex and insufficiently mature. Overall, although these methods are still actively developing and show promising performance, they largely follow a one-to-one paradigm and are unable to explicitly model uncertainty in the segmentation process, which is crucial in medical applications.

\noindent\textbf{Deep Generative Models} were initially developed for image generation and synthesis, but their ability to learn underlying data distributions has recently drawn increasing attention in medical image segmentation and uncertainty modeling. Among them, VAEs~\cite{vae}, which rely on probabilistic latent variable modeling, and GANs~\cite{gan}, which learn through adversarial training, laid the foundation for modern deep generative modeling. More recently, DMs~\cite{ddpm}, which generate images by gradually adding and then removing noise through a Markov chain process, have achieved state-of-the-art performance across various tasks, often matching or surpassing advanced GAN techniques~\cite{diffusion_beat_gan}. By incorporating conditioning into the generation process, several diffusion-based models have been successfully applied to various domains, including medical image segmentation. For example, EnsembleDiff~\cite{ensemblediff} concatenates medical images with their noisy counterparts during denoising, whereas SegDiff~\cite{segdiff} instead injects image features via element-wise addition after convolutional encoding. Notably, MedSegDiff~\cite{medsegdiff} and MedSegDiff-v2~\cite{medsegdiffv2} further introduce a Feature Frequency Parser to mitigate high-frequency noise and employ a Spectrum-Space Transformer to enhance semantic feature interaction. Other conditioning approaches include guided diffusion for universal anatomical segmentation~\cite{salsi2025guided} and prompt-guided variants such as ProGiDiff~\cite{progidiff}, which adapt pre-trained latent DM for flexible conditioning. In parallel, DMs have evolved to reduce computational cost along two main directions: performing the diffusion process in a low-dimensional latent space, as exemplified by Stable Diffusion~\cite{ldm}, and accelerating inference through improved sampling strategies such as DDIM~\cite{ddim} and PLMS~\cite{liu2022pseudo}. Latent diffusion has demonstrated strong efficiency across various generative tasks~\cite{ldm}, while accelerated sampling has been widely explored in medical image segmentation. For example, LSegDiff~\cite{lsegdiff} and LDSeg~\cite{ldseg} propose latent diffusion frameworks that achieve faster sampling and reduced memory consumption compared to pixel-space methods.

While prior studies highlight the efficiency of latent diffusion for medical image segmentation, most rely on standard VAEs or pixel-space conditioning and do not explicitly address tiny or sparse targets through specialized reconstruction losses. To overcome this limitation, we propose MedSegLatDiff, a latent diffusion framework that integrates VQ-VAE~\cite{vqvae,vqvae2} for stable compression of images and masks, and replaces MSE with a WCE loss to better preserve small and sparse structures. Moreover, stochastic latent sampling enables a one-to-many paradigm that captures inter-expert variability, making the model well suited for ambiguous cases and uncertainty-aware clinical decision-making.

\section{Method} \label{sec:method}

\begin{figure*}[t!]
    \vspace{-0.5cm}
    \centering
    \includegraphics[width=\textwidth]{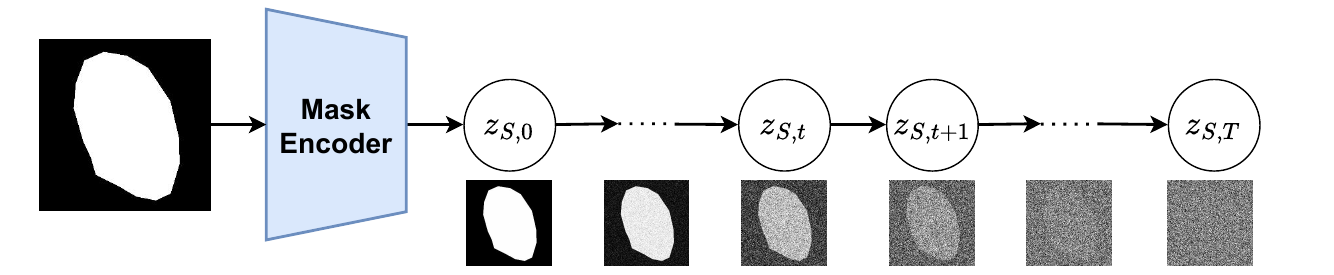}
    \caption{Illustration of the forward diffusion process in latent space, in which Gaussian noise is progressively added to the encoded segmentation mask across diffusion timesteps.}
    \label{fig:forward-process}
    \vspace{-0.25cm}
\end{figure*}

In this section, we introduce our proposed MedSegLatDiff framework, which integrates a conditional diffusion-based model with two VAEs for medical image segmentation. We begin with an overview of conditional diffusion models for image-space segmentation, followed by a description of the VAEs used to encode and decode medical images and segmentation masks. We then present our latent diffusion model with conditioning mechanisms operating in the latent space for this task.

\subsection{Segmentation in image space} \label{subsec:dm}

Medical image segmentation is formulated under the assumption that a labeled dataset is available, where each image is paired with a ground-truth segmentation mask. Formally, we consider a medical imaging dataset:
\begin{equation}
    \mathcal{D} = \{(X^{(i)}, S^{(i)})\}_{i=1}^{N},
\end{equation}
where \(X^{(i)}\in\mathbb{R}^{H \times W \times C}\) denotes the \(i\)-th medical image, \(S^{(i)}\in\{0, 1\}^{H \times W}\) is its corresponding segmentation mask, and \(N\) is the total number of image–mask pairs in this dataset \(\mathcal{D}\). The main idea when applying DM for the segmentation task is to gradually add noise \(\epsilon \sim \mathcal{N}(0, I)\) to a mask \(S\) through a forward process and progressively remove noise from the noisy image via a reverse process. The forward process starts with the original image \(S_0\) sampled from the mask distribution \(p_{\textrm{mask}}(S)\) and is defined as:
\begin{equation}
    q(S_t \mid S_{t-1}) = \mathcal{N}(S_t; \sqrt{\alpha_t} S_{t-1}, (1 - \alpha_t)I),
    \label{eq:forward}
\end{equation}
\(S_t\) denotes the latent variable at step \(t\) in the Markov chain, and \(\alpha_t\) is the noise scheduler controlling the noise variance at each step. Over the course of \(T\) steps, the forward process generates progressively noisier images \((S_1, S_2, \dotsc, S_T)\), with \(S_T \sim \mathcal{N}(0, I)\) representing the noisiest image. Given \(S_0\), a desirable property of this process \cite{ddpm} is that the marginal distribution of \(S_t\) can be obtained by analytically marginalizing out the intermediate latent variables:
\begin{equation}
    q(S_t \mid S_0) = \mathcal{N}(S_t; \sqrt{\gamma_t} S_0, (1 - \gamma_t)I).
    \label{eq:nice-property}
\end{equation}
where \(\gamma_t = \prod_{i=1}^t \alpha_i\). This forward process generates noisy image pairs that serve as training data for a denoising model \(\epsilon_{\theta}(S_t, t)\). The model takes the noisy image \(S_t\) as input and the added noise \(\boldsymbol{\epsilon}\) at step \(t\) as the target. In its simplest form, the model is trained using the MSE loss:
\begin{equation}
    \mathcal{L}_{\textrm{simple}} = \mathbb{E}_{\epsilon \sim \mathcal{N}(0,I), t} \left\| \epsilon - \epsilon_{\theta}(S_t, t) \right\|_2^2,
    \label{eq:dm-loss}
\end{equation}
and the noise estimator $\epsilon_{\theta}(S_t, t)$ is implemented using a U-Net architecture~\cite{ddpm}.

During inference, the reverse distribution $q(S_{t-1} \mid S_t, S_0)$ is approximated using the trained denoising model. Starting from pure Gaussian noise $S_T \sim \mathcal{N}(0, I)$, the noise is iteratively removed at each timestep to recover $S_{t-1}$. This procedure is referred to as the reverse diffusion process and is performed as follows:
\begin{equation}
    S_{t-1} = \frac{1}{\sqrt{\alpha_t}} \left( S_t - \frac{1 - \alpha_t}{\sqrt{1 - \gamma_t}} \epsilon_{\theta}(S_t, t) \right) + \sqrt{1-\alpha_t} \epsilon'
    \label{eq:reverse},
\end{equation}
where \(\epsilon' \sim \mathcal{N}(0, I)\) introduces stochasticity into the reverse process, and \(t = T, \dots, 1\) denotes the reverse diffusion steps. By repeating the reverse process for $T$ timesteps, the model generates a segmentation mask $S_0$ as an approximate sample from the underlying distribution of segmentation masks. Moreover, repeated stochastic sampling enables the generation of multiple plausible segmentation masks, effectively approximating this distribution.

However, the formulation described above only generates unconditional random masks and cannot perform segmentation conditioned on a given input image. To address this limitation, following prior studies, we incorporate conditioning into the generation process, enabling the model to generate a corresponding segmentation mask for a given input image. Specifically, given a medical image $X$ and its corresponding segmentation mask $S$, we adapt the model for the segmentation task by appending $X$ as an additional input channel to incorporate anatomical information into each noisy mask $S_t$. As a result, the noisy mask and the input image are concatenated channel-wise to obtain
\begin{equation}
    S_t^{\mathrm{cond}} := S_t \oplus X,
\end{equation}
where $\oplus$ denotes channel-wise concatenation. The resulting conditioned representation is then fed into the modified denoising model $\epsilon(S_t^{\mathrm{cond}}, t)$. This concatenation guides the denoising process, ensuring that the model generates new masks corresponding to the input image \(X\). During the forward process, noise is added only to the ground truth segmentation \(S\), while \(X\) remains unchanged to preserve anatomical guidance. Since the sampling process is stochastic, the diffusion model generates multiple segmentation masks \(S_0\) for each input image \(X\). When trained with a single segmentation mask per image, the model implicitly produces an ensemble of masks, effectively emulating the consensus-driven segmentation process of a group of doctors and potentially improving overall performance.

\subsection{Image and Mask in latent space} \label{subsec:vae}

The VAE~\cite{vae} is a neural network that maps each image from the data distribution to a latent code in a lower-dimensional latent space. It consists of an encoder \(\mathcal{E}(\cdot)\) and a decoder \(\mathcal{D}(\cdot)\). The encoder maps an image \(X \in \mathbb{R}^{H \times W \times C}\) to a latent representation \(z = \mathcal{E}(X)\), while the decoder reconstructs the image as \(X' = \mathcal{D}(z) = \mathcal{D}(\mathcal{E}(X))\), where \(z \in \mathbb{R}^{h \times w \times c}\) lies in a lower-dimensional latent space. The encoder downsamples the image by a factor \(f = \frac{H}{h} = \frac{W}{w}\), and in our study, we investigate different downsampling factors of the form \(f = 2^m\) with \(m \in \mathbb{N}\). For our segmentation task, both the medical image \(X\) and the corresponding segmentation mask \(S\) must be transformed into the same low-dimensional latent space. To achieve this, we employ two separate autoencoders trained independently for images and masks. In particular, we adopt the Vector Quantized Variational Autoencoder (VQ-VAE) \cite{vqvae, vqvae2}, whose discrete latent space has demonstrated strong generative performance in recent models such as Stable Diffusion~\cite{ldm}. VQ-VAE is particularly suitable for our setting because: (i) its discrete codebook provides a stable, expressive latent representation; (ii) the quantized latent space is naturally compatible with diffusion models, as demonstrated in Stable Diffusion~\cite{ldm}; and (iii) the stop-gradient mechanism stabilizes training and preserves the piecewise-constant structure characteristic of segmentation masks. 

First, we employ a VQ-VAE to encode the input image into a latent space and guide the generation process, ensuring that the predicted mask corresponds to the input image rather than being arbitrary. It consists of an encoder \(\mathcal{E}_X(\cdot)\) that maps the input image \(X\) to a latent representation \(z_X\), and a decoder \(\mathcal{D}_X(\cdot)\) that reconstructs the image. In addition, a quantization module \(Q_{\mathcal{V}_X}(\cdot)\) is designed to discretize the continuous latent representation \(z_X\) into \(\bar{z_X}\) using a codebook \(\mathcal{V}_X = \{v_X^{(k)}\}_{k=1}^K\). Specifically, each latent embedding vector in \(z_X\) is replaced by its nearest codebook entry, and the codebook is jointly learned during training. The optimization objective combines three synergistic terms: a reconstruction loss to ensure high-fidelity reconstructions
\begin{equation}
    \mathcal{L}_{\mathrm{rec}} = \|X - \mathcal{D}_{X}(\bar z_X)|_2^2,
\end{equation}
a codebook loss to align encoder outputs with the nearest embeddings
\begin{equation}
    \mathcal{L}_{\mathrm{codebook}} = \|\mathrm{sg}(z_X) - \bar z_X\|_2^2,
\end{equation}
and a commitment loss to prevent the encoder outputs from fluctuating excessively
\begin{equation}
    \mathcal{L}_{\mathrm{commit}} = \beta \|z_X - \mathrm{sg}(\bar z_X)\|_2^2.
\end{equation}
Here, \(\text{sg}[\cdot]\) denotes the stop-gradient operator, which prevents gradients from flowing into its argument during backpropagation, ensuring that the encoder and the codebook are updated separately. The full objective is then given by
\begin{equation}
    \mathcal{L}_{\mathrm{VQ\text{-}VAE}} 
    = \mathcal{L}_{\mathrm{rec}} 
    + \mathcal{L}_{\mathrm{codebook}} 
    + \mathcal{L}_{\mathrm{commit}}.
\end{equation}

Secondly, we design another VQ-VAE for the segmentation mask to enable the diffusion-based model to learn the mask distribution in the latent space more efficiently and effectively. It is structured similarly to the image VQ-VAE and consists of an encoder \(\mathcal{E}_{S}(\cdot)\), a decoder \(\mathcal{D}_{S}(\cdot)\), and a quantization module \(Q_{\mathcal{V}_S}(\cdot)\), which discretizes the latent representation using a codebook \(\mathcal{V}_S = \{v_S^{(k)}\}_{k=1}^K\). The key difference lies in our modified loss function, which focuses on the segmented region, with particular emphasis on tiny nodules in the LIDC-IDRI dataset. Instead of using MSE for reconstruction, we adopt WCE to better capture the characteristics of small and sparse lesion regions. The modified reconstruction loss assigns larger weights to pixels within the segmentation mask, enabling the model to focus on these regions and reconstruct the mask more accurately, while all other loss components remain unchanged. This design prevents the perceptual compression model from ignoring tiny nodules, which might otherwise be mistakenly treated as noise.

\subsection{Segmentation in latent space} \label{subsec:ldm}

\begin{figure*}[htbp]
    \vspace{-0.5cm}
    \centering
    \includegraphics[width=\textwidth]{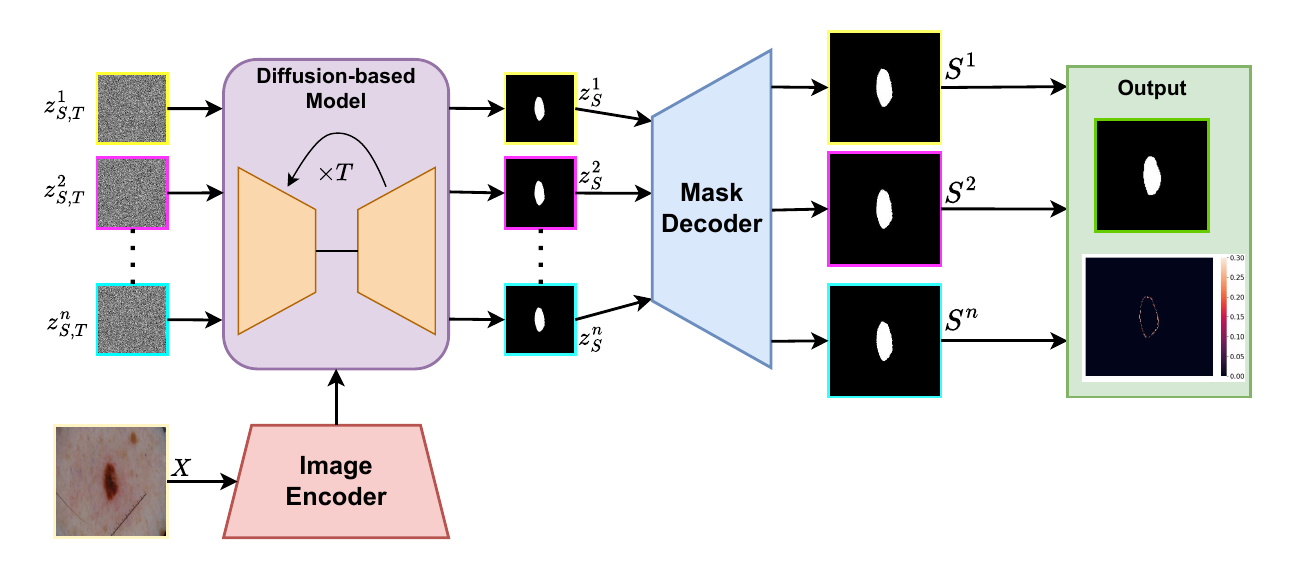}
    \caption{Overview of segmentation pipeline: multiple segmentation masks are generated from random noise in the latent space for a given medical image, followed by decoding, consensus fusion, and confidence map estimation.}
    \label{fig:reverse-process}
    \vspace{-0.25cm}
\end{figure*}

Once the two pre-trained VQ-VAE models, one for medical images and the other for the corresponding segmentation masks, are prepared as described in Part \ref{subsec:vae}, we perform the diffusion processes in the latent domain. For each pair of medical image \( X \in \mathbb{R}^{H \times W \times C} \) and 
segmentation mask \( S \in \mathbb{R}^{H \times W} \), the corresponding latent 
representations are obtained as \(\bar{z}_X\) and \(\bar{z}_S\). Using the latent representations, the diffusion model performs the forward diffusion process on each latent mask \(z_{S,0}=\bar z_S\), as illustrated in Fig.~\ref{fig:forward-process}. This process follows the diffusion mechanism described in Eqs.~\ref{eq:forward} and~\ref{eq:nice-property}, which generate noisy labels for the data and train the denoising model. 

Similarly, to generate segmentation masks for a medical image \(X\) that are consistent with the input rather than arbitrary, the denoising model incorporates additional conditions derived from the image. These conditions guide the noise removal process and the generation of the corresponding segmentation mask. Specifically, the denoising model takes as input a combination of the noisy mask \(z_{S,t}\) and the latent image representation \(\bar{z}_X\) as a conditioning signal, using a concatenation operator in the latent space:  
\begin{equation}
    z_{S,t}^{\textrm{cond}} = z_{S,t} \oplus \bar{z}_X
    \label{concat}
\end{equation}
The denoising model \(\epsilon_{\theta}(z_{S,t}^{\textrm{cond}}, t)\) is then trained for the reverse diffusion process. Accordingly, the reverse process in Eq.~\ref{eq:reverse} is modified to incorporate the conditioning as follows:  
\begin{equation}
    S_{t-1} = \frac{1}{\sqrt{\alpha_t}} \left( S_t - \frac{1 - \alpha_t}{\sqrt{1 - \gamma_t}} \epsilon_{\theta}(z_{S,t}^{\textrm{cond}}, t) \right) + \sqrt{1-\alpha_t} \epsilon'
    \label{eq:reverse-modify}
\end{equation}
Thus, during the inference process, for each latent medical image \(\bar{z}_X\), a set of \(n\) segmentation masks in the latent space, \(\mathcal{Z}_{S,0} = \{z_S^1, z_S^2, \dots, z_S^n\}\), is generated from a corresponding set of Gaussian noises, \(\mathcal{Z}_{S,T} = \{z_{S,T}^1, \dots, z_{S,T}^n\}\). The final step is to map the latent masks back to the original image space using the segmentation mask decoder:  
\begin{equation}
    S^i = \mathcal{D}_S(z_S^i),
\end{equation}
where \(i=1,2,\dots,n\). The final output of the model is the set of segmentation masks \(\mathcal{S} = \{S^1, S^2, \dots, S^n\}\), which can be used to compute a confidence map and generate an implicit ensemble segmentation map. The final consistent result is obtained by averaging the outputs in the set to compute a confidence map, which is then thresholded at 0.5 to produce the final binary segmentation mask. The entire segmentation process is illustrated in Fig.~\ref{fig:reverse-process}. It can fully mimic the segmentation performed by a group of \(n\) doctors, which helps to improve the model's performance.

\section{Experiments} \label{sec:experiments}

\subsection{Datasets} \label{subsec:dataset}

We conducted experiments on three publicly available datasets. The first dataset, ISIC-2018~\cite{isic2018}, was released for the ISIC Skin Image Analysis Workshop and Challenge 2018 (Task 1), which focuses on automated binary segmentation of skin lesion regions in dermoscopic images. The dataset was obtained from the official ISIC Challenge website (\url{https://challenge.isic-archive.com/data/#2018}) and consists of dermoscopic images collected from multiple clinical centers. The second dataset, CVC-Clinic~\cite{cvc-clinic}, comprises frames extracted from endoscopy videos and is widely used for polyp segmentation in medical images. It is publicly available on the Kaggle repository 
(\url{https://www.kaggle.com/datasets/balraj98/cvcclinicdb}). The third dataset, LIDC-IDRI~\cite{lidc-idri}, consists of thoracic CT scans with lesion annotations provided by four experienced radiologists. Only slices containing nodules were extracted, and the four radiologist-annotated masks were averaged to form a single consensus mask used as the ground-truth target. Unlike the first two datasets, LIDC-IDRI was split on a patient-wise basis into training, validation, and testing sets with an 8:1:1 ratio. For ISIC-2018 and CVC-Clinic, the default dataset-provided splits were used.

\subsection{Implementation Details} \label{subsec:implementation_detail}

We employ a VQ-VAE to compress both medical images and segmentation masks, followed by a diffusion-based model operating in the latent space for segmentation. The VQ-VAE adopts an autoencoder architecture with three downsampling and upsampling levels, while the diffusion model is implemented as a U-Net with a matching three-level configuration. Both networks use 64 base channels with channel multipliers $[1, 2, 4]$. For mask compression, a Weighted Cross-Entropy (WCE) loss is applied, with a positive-class weight of 5 for ISIC-2018 and CVC-Clinic, and 50 for LIDC-IDRI to account for the extremely small target regions. The diffusion model follows the training setup described in~\cite{ddpm}, using 1000 diffusion timesteps. To condition the diffusion process, the encoded input image is directly concatenated with the noisy latent representation, serving as guidance during both training and inference. Images from ISIC-2018 and CVC-Clinic are resized to $256 \times 256$ pixels, while LIDC-IDRI slices are resized to $128 \times 128$. Training is performed using the AdamW optimizer~\cite{adam}, with batch sizes of 32 for ISIC-2018 and CVC-Clinic and 64 for LIDC-IDRI. The initial learning rate is set to $1 \times 10^{-4}$. During inference, five stochastic samples are generated per input and aggregated by averaging to obtain a confidence map, which is thresholded at 0.5 to produce the final segmentation. For fair comparison, diffusion-based baselines are reproduced under identical settings. Performance is evaluated using PSNR~\cite{huynh2008scope} and SSIM~\cite{wang2004image} for the reconstruction task, and the Dice coefficient and Intersection over Union (IoU) for the segmentation task. All experiments are implemented in PyTorch and conducted on four Tesla V100 GPUs with 16~GB memory each.

\subsection{Results} \label{subsec:main_results}

We first trained two separate VQ-VAE models for images and segmentation masks to enable latent-space mapping, and then trained a conditional diffusion model on the learned latent distributions. Table~\ref{tab:result_vae} reports the reconstruction performance of the VQ-VAE models

\begin{table}[htbp]
    \centering
    \caption{Performance of VQ-VAE for perceptual data compression.}
    \label{tab:result_vae}
    \resizebox{\columnwidth}{!}{
        \begin{tabular}{llcccc}
            \toprule
            Dataset & Input (loss) & Dice & IoU & SSIM & PSNR \\
            \midrule
            \multirow{3}{*}{ISIC-2018}
                & Mask (MSE) & 98.0 & 96.5 & 0.975 & 38.0 \\
                & Mask (WCE) & 98.4 & 96.9 & 0.98 & 38.4 \\
                & Image (MSE) & -- & -- & 0.94 & 34.2 \\
            \midrule
            \multirow{3}{*}{CVC-Clinic}
                & Mask (MSE) & 99.0 & 98.9 & 0.976 & 38.0 \\
                & Mask (WCE) & 99.5 & 99.0 & 0.98 & 38.7 \\
                & Image (MSE) & -- & -- & 0.94 & 34.5 \\
            \midrule
            \multirow{3}{*}{LIDC-IDRI}
                & Mask (MSE) & 88.0 & 83.1 & 0.92 & 35.2 \\
                & Mask (WCE) & 94.4 & 89.4 & 0.99 & 41.7 \\
                & Image (MSE) & -- & -- & 0.89 & 33.2 \\
            \bottomrule
        \end{tabular}
    }
\end{table}

For image reconstruction, we employed the mean squared error (MSE) loss, achieving high reconstruction quality with SSIM of 0.94 and PSNR of approximately 34 on the ISIC-2018 and CVC-Clinic datasets, while lower performance was observed on the more challenging LIDC-IDRI dataset (SSIM 0.89, PSNR 33.2). For mask compression, using MSE loss resulted in strong segmentation fidelity, with Dice scores exceeding 98\% and IoU scores above 96\% on ISIC-2018 and CVC-Clinic. However, the presence of small and fine-grained structures in LIDC-IDRI led to reduced performance, with a Dice score of 88.0\%. When switching to the WCE loss, performance improvements on the ISIC-2018 and CVC-Clinic datasets are marginal, as their segmentation masks typically cover relatively large regions. In contrast, the LIDC-IDRI dataset shows substantial improvements, with Dice increasing by 6.4 points to 94.4, IoU improving by 6.3 points to 89.4, SSIM increasing by 0.07, and PSNR improving by 6.5 dB. Overall, only modest gains are observed on datasets with larger mask regions, while WCE proves particularly beneficial for datasets containing small and sparse structures. These results validate VQ-VAE's effectiveness for latent-space compression of medical data, especially with WCE for sparse/tiny masks.

\begin{figure}[htbp]
    \vspace{-0.5cm}
    \centering
    \includegraphics[width=\columnwidth]{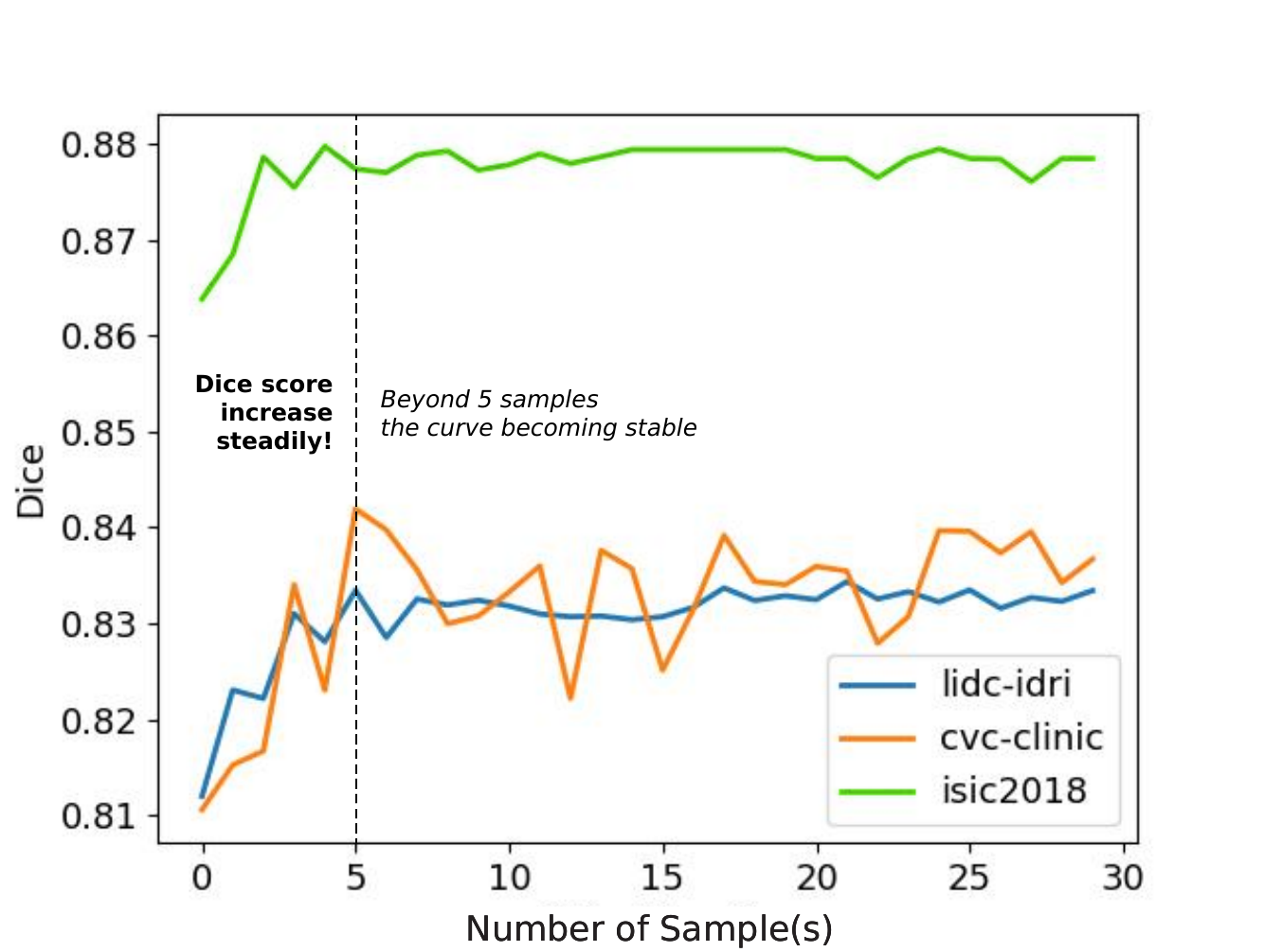}
    \caption{Dice scores versus the number of samples}
    \label{fig:n_samples}
    \vspace{-0.5cm}
\end{figure}

\begin{figure*}[t!]
    \vspace{-0.5cm}
    \centering
    \includegraphics[width=\textwidth]{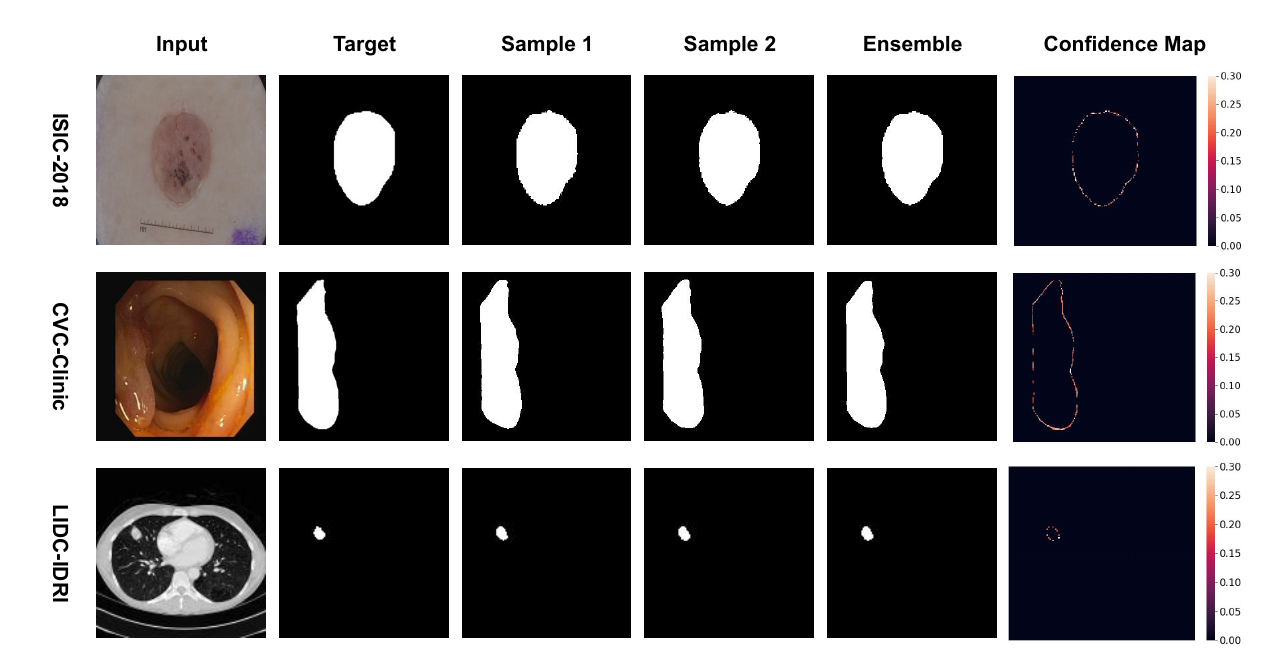}
    \caption{Visualization of multiple segmentation masks. The first two columns show the input image and the ground-truth mask, followed by two mask samples, their averaged output, and the resulting confidence map.}
    \label{fig:ensemble}
    \vspace{-0.25cm}
\end{figure*}

For the segmentation task, qualitative results of our model are first presented in Fig.~\ref{fig:ensemble}. Five stochastic samples are generated for each input image and averaged to produce a confidence map, which is then thresholded at 0.5 to obtain the final segmentation mask. Results across all three datasets demonstrate that the generated masks closely match the ground-truth annotations, including the small target regions in the LIDC-IDRI dataset. Variations among sampled masks are mainly observed along object boundaries, which can be attributed to inherent ambiguities in medical images. By generating multiple segmentation masks, the proposed model produces a confident consensus prediction while explicitly capturing uncertainty through the confidence map. This uncertainty-aware output enhances the interpretability of the model and provides clinicians with additional insights for more informed analysis. Furthermore, we present the Dice score as a function of the number of samples in Fig.~\ref{fig:n_samples} to justify our choice of using five samples during inference. The segmentation performance improves steadily as the number of samples increases from 1 to 5, with Dice scores on ISIC-2018 rising from 0.865 to 0.880, on CVC-Clinic from 0.810 to 0.845, and on LIDC-IDRI from lower values to 0.834, indicating improved consensus as more stochastic samples are aggregated. Beyond five samples, performance gains gradually plateau (up to 30 samples), while computational cost continues to increase. Therefore, we adopt five samples as an effective trade-off between segmentation accuracy and inference efficiency.

\begin{table}[htbp]
    \centering
    \caption{Comparison of MedSegLatDiff with previous one-to-one segmentation models.}
    \label{tab:compare_1-1}
    \resizebox{\columnwidth}{!}{
        \begin{tabular}{lcccccc}
            \toprule
            & \multicolumn{2}{c}{ISIC-2018}
            & \multicolumn{2}{c}{CVC-Clinic}
            & \multicolumn{2}{c}{LIDC-IDRI} \\
            \cmidrule(lr){2-3} \cmidrule(lr){4-5} \cmidrule(lr){6-7}
            Model & Dice & IoU & Dice & IoU & Dice & IoU \\
            \midrule
            UNet~\cite{unet} & 85.6 & 78.5 & 82.3 & 72.5 & 81.0 & 70.0 \\
            UNet++~\cite{unet++} & 81.0 & 72.9 & 79.4 & 70.9 & 79.9 & 68.9 \\
            ResUNet~\cite{resunet} & 87.1 & 78.2 & 81.5 & 71.6 & 82.1 & 70.6 \\
            nnUNet~\cite{nnunet} & 87.5 & 79.5 & 81.3 & 71.3 & 82.8 & 71.0 \\
            \midrule
            MedSegLatDiff & \textbf{88.0} & \textbf{80.5} & \textbf{84.5} & \textbf{73.1} & \textbf{83.4} & \textbf{71.8} \\
            \bottomrule
        \end{tabular}
    }
\end{table}

Moreover, quantitative comparisons between our method and existing approaches are reported in Tables~\ref{tab:compare_1-1} and~\ref{tab:compare_1-n}. Table~\ref{tab:compare_1-1} shows that MedSegLatDiff achieves competitive performance compared with one-to-one baselines. In particular, MedSegLatDiff obtains the highest Dice scores (88.0 / 84.5 / 83.4) and IoU (80.5 / 73.1 / 71.8) across all datasets, with consistent improvements ranging from 0.5 to 2.2 in Dice and from 0.5 to 0.8 in IoU compared to the best-performing competing methods. These results indicate the advantage of the proposed one-to-many modeling paradigm over conventional one-to-one approaches, highlighting its suitability for medical image segmentation tasks where ambiguity and inter-observer variability are inherent. In parallel, comparisons with one-to-many diffusion-based baselines are reported in Table~\ref{tab:compare_1-n}. MedSegLatDiff demonstrates competitive performance against other diffusion-based methods. Specifically, while it achieves comparable IoU and slightly lower Dice than MedSegDiff on ISIC-2018, it clearly excels on CVC-Clinic, improving Dice by 0.5 and IoU by 0.6. On the more challenging LIDC-IDRI dataset, where tiny and sparse masks pose greater difficulties, MedSegLatDiff further improves Dice by 0.9 and IoU by 0.7, highlighting its robustness in handling small target structures. Overall, these results demonstrate the effectiveness of latent-space diffusion with VQ-VAE compression and WCE in achieving robust and uncertainty-aware medical image segmentation.

\begin{table}[htbp]
    \centering
    \caption{Comparison of MedSegLatDiff with recent one-to-many segmentation models.}
    \label{tab:compare_1-n}
    \resizebox{\columnwidth}{!}{
        \begin{tabular}{lcccccc}
            \toprule
            & \multicolumn{2}{c}{ISIC-2018}
            & \multicolumn{2}{c}{CVC-Clinic}
            & \multicolumn{2}{c}{LIDC-IDRI} \\
            \cmidrule(lr){2-3} \cmidrule(lr){4-5} \cmidrule(lr){6-7}
            Model & Dice & IoU & Dice & IoU & Dice & IoU \\
            \midrule
            EnsembleDiff~\cite{ensemblediff} & 86.6 & 79.5 & 83.5 & 71.9 & 81.7 & 70.5 \\
            SegDiff~\cite{segdiff} & 85.3 & 78.7 & 82.1 & 70.1 & 81.0 & 70.0 \\
            MedSegDiff~\cite{medsegdiff} & \textbf{88.1} & \textbf{80.5} & 84.0 & 72.5 & 82.5 & 71.1 \\
            \midrule
            MedSegLatDiff & 88.0 & \textbf{80.5} & \textbf{84.5} & \textbf{73.1} & \textbf{83.4} & \textbf{71.8} \\
            \bottomrule
        \end{tabular}
    }
\end{table}

\noindent\textbf{Limitation: } Despite its promising performance, our method has several limitations. We currently employ VQ-VAE for latent-space segmentation without systematically comparing different VAE variants or analyzing their invariance properties. Although WCE outperforms MSE for mask compression, further investigation of loss functions tailored to tiny or sparse structures (e.g., Focal or Dice-based losses) is needed. These directions will be explored in future work to further enhance robustness and uncertainty modeling.

\section{Conclusion} \label{sec:conclusion}
In this paper, we propose MedSegLatDiff, a conditional diffusion framework operating in a compressed latent space via dual VQ-VAE modules for images and segmentation masks, enabling efficient perceptual compression while preserving critical structural details. By replacing the conventional MSE loss with WCE loss for mask reconstruction, the model significantly improves the recovery of tiny and sparse structures, such as lung nodules in LIDC-IDRI. Experiments show that generating five stochastic samples achieves a strong trade-off between computational cost and segmentation accuracy, producing stable ensemble predictions and reliable confidence maps. MedSegLatDiff consistently outperforms traditional one-to-one models and achieves competitive or superior performance compared to recent one-to-many diffusion-based approaches on ISIC-2018, CVC-Clinic, and LIDC-IDRI. The latent-space design suppresses noise and enables the diffusion model to focus on segmentation, while the one-to-many paradigm captures uncertainty by mimicking inter-expert variability, thereby supporting clinical decision-making.

For future work, we plan to investigate advanced conditioning strategies, such as classifier guidance~\cite{diffusion_beat_gan}, classifier-free guidance~\cite{classifier-free}, and flow-based generative modeling, to further enhance output stability, calibration, and uncertainty quantification.

\bibliographystyle{IEEEtran}

\end{document}